\documentclass{article}
\usepackage{iclr2027_conference,times}
\usepackage[T1]{fontenc}
\usepackage{amsmath,amsfonts,bm}

\def\eqref#1{equation~\ref{#1}}
\def\1{\bm{1}}

\DeclareMathAlphabet{\mathsfit}{\encodingdefault}{\sfdefault}{m}{sl}
\SetMathAlphabet{\mathsfit}{bold}{\encodingdefault}{\sfdefault}{bx}{n}

\usepackage{hyperref}
\usepackage{url}
\usepackage{booktabs}
\usepackage{amsmath}
\usepackage{amssymb}
\usepackage{graphicx}
\usepackage{tabularx}
\usepackage{array}
\usepackage{listings}
\usepackage{xcolor}

\newcolumntype{Y}{>{\raggedright\arraybackslash}X}
\newcolumntype{R}{>{\raggedleft\arraybackslash}X}
\newcolumntype{d}[1]{>{\raggedleft\arraybackslash}p{#1}}

\makeatletter
\def\paragraph{\@startsection{paragraph}{4}{\z@}%
  {0.4ex plus 0.12ex minus .1ex}{-0.4em}{\normalsize\bf}}
\def\section{\@startsection {section}{1}{\z@}{-2.0ex plus
    -0.5ex minus -.2ex}{1.1ex plus 0.2ex
minus0.15ex}{\large\sc\raggedright}}
\def\subsection{\@startsection{subsection}{2}{\z@}{-1.8ex plus
-0.5ex minus -.2ex}{0.6ex plus .15ex}{\normalsize\sc\raggedright}}
\def\subsubsection{\@startsection{subsubsection}{3}{\z@}{-1.5ex
plus      -0.5ex minus -.2ex}{0.35ex plus
.15ex}{\normalsize\sc\raggedright}}
\makeatother

\newcommand{\Slp}{S^{\mathrm{lp}}_{\theta}}
\newcommand{\Sjsd}{S^{\mathrm{js}}_{\theta}}
\newcommand{\Sflip}{S^{\mathrm{fl}}_{\theta}}

\newcommand{\bench}{DECOY}

\title{\fontsize{15}{18}\selectfont\raggedright Confidence Without Cause: Predicting and Localizing Physics-Shortcut Sensitivity in Language Model Hidden States\par}

\author{{\fontsize{11}{13}\selectfont\bfseries
Shubham Mishra\textsuperscript{$\spadesuit$}\quad
Atharv Sharma\textsuperscript{$\spadesuit$}\quad
Dhruv Kumar\textsuperscript{$\spadesuit$}}\\[8pt]
\textsuperscript{$\spadesuit$}Birla Institute of Technology and Science, Pilani\\[6pt]
\textbf{Correspondence:} \href{mailto:f20220763@pilani.bits-pilani.ac.in}{\texttt{f20220763@pilani.bits-pilani.ac.in}}}

\usepackage{etoolbox}
\makeatletter
\patchcmd{\@maketitle}
  {Anonymous authors\\Paper under double-blind review}
  {\@author}
  {}{\PackageError{physmon}{Could not replace anonymous author block}{}}
\makeatother

\renewcommand{\iclrruler}[1]{}

\begin{document}
\maketitle
\begin{abstract}
A language model's confidence on a physics problem deserves trust only if it ignores details irrelevant to the answer. We introduce DECOY (Distractor-Embedded Certified cOunterfactuals for phYsics): 155 families of four prompts differing in one irrelevant detail. A symbolic solver certifies that each edit preserves the answer. Families are labelled by the largest drop in the model's log-probability of the certified answer, using a threshold fixed before probe training. Qwen2.5-7B-Instruct is sensitive on 50.0\% of the 140 families yielding labels, and Llama-3.1-8B-Instruct on 22.9\%. Effects are largest when distractor values lie numerically close to the answer. A linear probe on Qwen's layer-18 hidden state, averaged over four renderings, predicts sensitivity before generation at AUROC 0.760 with whole families held out. A separately tuned single-rendering probe reaches 0.747. A frozen layer-16 probe reaches 0.820 pooled AUROC on 15 families built after training, though within-cue-type AUROC is lower (0.667-0.6875). Under a paired family bootstrap, the probe performs on par with a fitted token-entropy baseline (AUROC 0.727) and achieves the lowest Brier score among tested predictors (0.204 versus 0.214). Knocking out four attention heads at layer 16 cancels sensitivity on a 20-family panel, localising it to a narrow mechanism. These results show that Qwen's confidence can track irrelevant details, that this failure is detectable before generation, and that targeted intervention can suppress it. DECOY provides certified ground truth and paired-tested baselines for evaluating hidden-state monitors.
\end{abstract}

\section{Introduction}

Consider the problem \emph{``Object A has mass 2.0 kg and moves at 6.0 m/s. The laboratory floor area
is 10 m\textsuperscript{2}. What is the kinetic energy of object A?''} The floor area appears in no
equation a physicist would write, and a symbolic check of the governing expression confirms it.
Qwen2.5-7B-Instruct answers 36.0 J, correctly, on this prompt and on three copies in which the floor
area is 20, 40 and 80 m\textsuperscript{2}. Its confidence moves all the same. The log-probability it
assigns to the correct answer falls by 1.03 nats between the first and the last copy. A nat is the
unit of natural-log probability, and a drop of 0.5 nats means the probability falls by a factor of about
1.65. Across the benchmark introduced here, this model shows a drop of at least 0.5 nats on half
of all labelled problem families.

A model whose confidence depends on details that the physics says to ignore is a reliability
concern, and the concern is sharpest when such dependence can be found and flagged before an answer
is used. Two things stand in the way. The first is measurement. Benchmarks built by perturbing
existing problems \citep{shi2023largelanguagemodelseasily,
mirzadeh2025gsmsymbolicunderstandinglimitationsmathematical} rely on template or heuristic
construction rather than a formal proof that a perturbation preserves the correct answer. As a
result, a change in output can still mix instability in the model with an unverified change in the task.
The second is cost. Sensitivity is a property of a set of prompts, and observing it takes
several queries per problem. We address both. \textbf{\bench{}} contains 155 families, each made of
four renderings of one physics problem that differ only in a designated \emph{cue}, a detail that
plays no role in the answer. For every family a computer algebra system \citep{Meurer_2017} proves
that the cue symbol is absent from the solved expression for the answer, so any change in the model's
behaviour across the family belongs to the model. The certificate also gives the measurement a
clear reference point. A model that answers by identifying the governing law and the quantities it
depends on computes the same thing on all four prompts, so its probability for the certified answer
is the same on all four. The size of the largest drop across the family therefore measures how far
the cue reaches into the model's computation, and it is zero for an ideal solver. We then ask whether
the model's own hidden state on a single prompt already predicts that drop, which would let a
deployed system flag a risky question with one forward pass and before any answer is produced.

It does. A linear probe on the hidden state (the residual stream) at layer 18 of Qwen's 28 layers, at the last prompt
token, predicts the family's label, averaged over a family's four renderings, at AUROC 0.760 with
whole families held out. Scored on a single rendering instead, a separately tuned probe reaches 0.747. It
exceeds a classifier on the prompt's words by 0.11 and holds at 0.738 after the part explained by a
correctness probe is removed. A layer-16 probe with frozen weights reaches 0.820 pooled on fifteen
families built after training, though within-cue-type AUROC is lower (Section~\ref{sec:monitor}).

A monitor earns its place by adding to what cheaper signals already provide, so we compare it with
eight non-activation predictors on the same families under a \emph{paired} family bootstrap. In this
bootstrap, both predictors are scored on identical resamples and the interval is taken on their
difference. The
probe outperforms every text-based and behavioural predictor, from a chain-of-thought classifier
(AUROC 0.599) to an LLM judge (AUROC 0.480), and performs on par with a \emph{fitted} token-entropy
baseline (AUROC 0.727), a classifier trained on the model's output uncertainty. It also reaches the
lowest Brier score of any single predictor. This comparison is stricter than the random-direction and
surface baselines common in probing work, and the closeness of the entropy baseline is itself a
useful result. It tells practitioners which cheap signal to deploy alongside a probe. Finally,
knocking out four attention heads at layer 16 of
Qwen cancels the sensitivity on a 20-family panel. The mean normalised recovery is 1.870 (recovery
above 1.0 reflects an over-reversal of the drop, and capping each family at 1.0 gives 0.969), against
0.021 for random perturbations of the same size.

\textbf{Contributions.} \textbf{(a) A certified benchmark:} \bench{}, 155 solver-certified
counterfactual physics families over three cue types and three domains, with stored certificates and
a three-level framework that keeps prompt condition, behaviour and internal state apart
(Section~\ref{sec:method}). \textbf{(b) A measurement of sensitivity:} under a pre-registered
threshold, sensitivity is common, differs sharply between models, and peaks when a distractor
nearly matches the answer (Section~\ref{sec:behaviour}). \textbf{(c) A pre-generation monitor:}
prompt-side hidden states predict sensitivity with families held out, and a frozen probe retains
signal on new families built afterwards (Section~\ref{sec:monitor}). \textbf{(d) An evaluation standard:}
a paired comparison against fitted baselines that any future hidden-state monitor can be checked
against (Section~\ref{sec:compare}). \textbf{(e) A causal handle in Qwen:} a localised set of attention heads
whose removal cancels the behaviour (Section~\ref{sec:causal}).

\begin{figure}[t]
\centering
\includegraphics[width=0.78\linewidth]{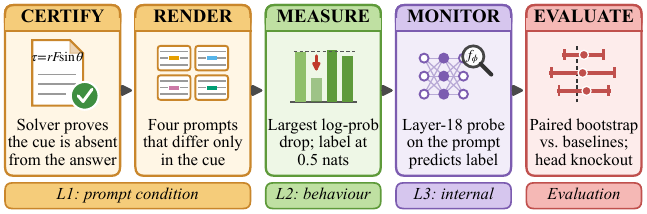}
\caption{Overview of the pipeline, following the three levels of Section~\ref{sec:framework}. A
solver certifies that the cue cannot affect the answer (Level 1). One teacher-forced pass per
rendering gives the family's sensitivity label (Level 2). A probe on the layer-18 hidden state of
Qwen2.5-7B-Instruct predicts that label before generation (Level 3), and the probe is evaluated
against non-activation baselines and by head knockout.}
\label{fig:system}
\end{figure}

\section{Related Work}\label{sec:related}

\paragraph{Shortcut learning and perturbation benchmarks.} Models often reach high accuracy through
features unrelated to the intended task \citep{Geirhos_2020}. In NLP this shows up as syntactic
heuristics \citep{McCoy_2019}. Contrast sets probe a model's local decision boundary with minimal
edits that typically change the gold label \citep{Gardner_2020}, and the invariance tests of
CheckList apply edits that should leave the label unchanged \citep{Ribeiro_2020}. \emph{We follow
the invariance-test design and add a symbolic proof that each edit preserves the answer}, which
makes the measured change a property of the model. In mathematical reasoning, irrelevant context lowers
accuracy on grade-school word problems \citep{shi2023largelanguagemodelseasily} drawn from GSM8K
\citep{cobbe2021trainingverifierssolvemath} and regenerated grade-school templates reveal
accuracy variance under structure-preserving edits
\citep{mirzadeh2025gsmsymbolicunderstandinglimitationsmathematical}, and controlled distractor
injection isolates effects on reasoning-path selection and accuracy \citep{yang2025llmreasoningdistracted}. \emph{We measure confidence
shifts within a family whose answer is certified constant}, so a family can be sensitive while every
answer is correct. Physics and science benchmarks
\citep{wang2024scibenchevaluatingcollegelevelscientific,
xu2025ugphysicscomprehensivebenchmarkundergraduate, Zhang_2025}
supply the problem material. \emph{We build on physics because dimensional structure and closed-form
solutions make answer invariance provable}.

\paragraph{Faithfulness and monitoring.} A model's written reasoning does not always reflect the
features that drive its answer \citep{Turpin_2023, lanham2023measuringfaithfulnesschainofthoughtreasoning}.
The same holds for reasoning-trained models \citep{chen2025reasoningmodelsdontsay} and for
reasoning produced without adversarial prompting \citep{arcuschin2026chainofthoughtreasoningwildfaithful}.
Monitoring the chain of thought is a valuable safety opportunity that is also imperfect and fragile
\citep{korbak2025chainthoughtmonitorabilitynew}, which motivates complementary monitors that read the
model's internal state. \emph{We test that motivation directly: chain-of-thought and
answer-rationale classifiers run as baselines against a hidden-state probe on the same families,
on a label whose correctness is certified rather than assumed.}

\paragraph{Probing internal states.} Linear probes on intermediate representations
\citep{alain2018understandingintermediatelayersusing} are a standard tool, and interpreting them
requires separating what the probe learns from what the model encodes
\citep{Belinkov_2022, Hewitt_2019}. We adapt
the control-task logic of \citet{Hewitt_2019} to this setting. \emph{Our domain-generalisation check
tests the probe's signal against label-shuffled permutations (Appendix~\ref{app:additional}), and we
separately ask whether a non-activation predictor already reaches the same held-out score.} We also
score a generic correctness probe as a full comparator, since a problem that makes a model unstable
is often a hard problem.

\paragraph{Truthfulness, self-knowledge and uncertainty.} A closely related line reads truthfulness
and correctness from internal states. Contrast-consistent search recovers a truth direction without
labels \citep{burns2024discoveringlatentknowledgelanguage}. Supervised probes detect when a model is
producing a false statement \citep{Azaria_2023}, and true and false statements separate along linear
directions \citep{marks2024geometrytruthemergentlinear}. We instead read a direction for sensitivity
to a detail whose irrelevance is proven, not assumed. Models can also estimate the probability that
their own answers are correct \citep{kadavath2022languagemodelsmostlyknow}, and representation
engineering reads and steers such concepts at scale \citep{zou2025representationengineeringtopdownapproach}.
On the output side, semantic entropy detects hallucination from the distribution of sampled answers
\citep{Farquhar_2024}. Semantic entropy probes instead recover a sampling-based estimate from a
single generation \citep{kossen2024semanticentropyprobesrobust}, INSIDE reads cross-generation hidden-state covariance
to detect hallucination \citep{chen2024insidellmsinternalstates}, and perturbation-based
representational stability calibrates confidence directly
\citep{khanmohammadi2025calibratingllmconfidenceprobing}. These target correctness and hallucination
under free generation, and evaluating them on \bench{} is a natural extension of the comparator
ladder in Section~\ref{sec:compare}. \emph{We place a hidden-state probe, a
correctness probe and fitted output-uncertainty baselines side by side on the same families}, which
shows how much of what a probe reads is shared with signals that need no access to activations.

\paragraph{Causal localisation.} Activation patching and causal mediation analysis identify
components whose alteration changes behaviour \citep{Meng_2022,
vig2020causalmediationanalysisinterpreting, wang2022interpretabilitywildcircuitindirect, Conmy_2023},
with known interpretive pitfalls \citep{heimersheim2024useinterpretactivationpatching}, and
competing mechanisms have been traced for factual and counterfactual content \citep{Ortu_2024}.
\emph{We apply multi-head knockout to a family-level behavioural statistic}, the same quantity the
probe predicts, and report random-direction, random-site, donor and damage controls next to the
effect. Following \citet{heimersheim2024useinterpretactivationpatching}, we read a large recovery as
evidence about a component's contribution to the behaviour.

\section{Methodology}\label{sec:method}

\subsection{Three-level framework}\label{sec:framework}

We separate three levels of description (Figure~\ref{fig:system}). \textbf{Level 1} is the
\emph{prompt condition}: which value of the cue a prompt contains. \textbf{Level 2} is \emph{observed
behaviour}: how much the model's output moves across a family whose correct answer is certified
constant. \textbf{Level 3} is the \emph{internal state}: a prompt-side hidden state that predicts
Level 2, or a component whose removal changes it. Our claims follow the chain Level 3 $\to$ predicts $\to$
Level 2 $\to$ correlated with $\to$ Level 1. Level 2 is a statement about the model's output. Whether the
model consulted the cue as a reasoning step is a further question, and we keep the two apart.

\subsection{The \bench{} benchmark}\label{sec:bench}

\paragraph{Templates and families.} Each family starts from a symbolic template
$\tau = (r, z, a, g, y^{*})$: the physically relevant variables $r$ (mass and speed), the
cue $z$ (the floor area), the stated assumptions $a$, the governing law $g$
($E = \tfrac{1}{2}mv^2$), and the answer $y^{*}$ derived by a solver. The family is the set of four
prompts
\begin{equation}
\mathcal{F}_{\tau} = \bigl\{\, x_{\tau}(r, z_i, a) \,\bigr\}_{i=1}^{4},
\end{equation}
where $x_{\tau}(r, z_i, a)$ is the natural-language rendering carrying the $i$-th cue value $z_i$.
The first rendering is the \emph{base} rendering (variant 0 in the released files). The families are built so that
\begin{equation}\label{eq:inv}
y^{*}_{\tau}(r, z_i, a) = y^{*}_{\tau}(r, z_j, a) \qquad \text{for all } i, j,
\end{equation}
that is, the correct answer is the same in all four prompts.

\paragraph{Certification.} Equation~\ref{eq:inv} is proved for every family. Let
$E = \mathrm{solve}(g, y^{*})$ be the governing law rearranged to give the answer, and let
$\Sigma(E)$ be the set of variables that still appear in $E$ after the assumptions $a$ are
substituted. The certificate for cue symbol $c$ is the machine-checked statement
\begin{equation}\label{eq:cert}
c \notin \Sigma(E),
\end{equation}
and a symbol absent from $E$ cannot change its value. This free-symbol check applies to Cue A and
Cue B, where the cue is not part of the governing quantity. For Cue C the cue is the governing
quantity restated in an equivalent form, so invariance is instead established by an exact identity
between the transformed and untransformed expressions. Three further checks complete the certificate:
$E$ evaluated at the stated parameters must match the declared answer to a relative tolerance of
$10^{-6}$, every parameter must be finite and physically plausible, and the declared answer must
pass the same parser later applied to model output. Appendix~\ref{app:examples} reproduces a
complete template, its certificate, its four prompts and the model outputs for one family.

\paragraph{Cue types.} \textbf{Cue A} adds a physically irrelevant variable, such as a wall
temperature in a centripetal-force problem. \textbf{Cue B} adds a distractor with the same units as
the answer that the stated setup makes irrelevant, such as the torque of a second, mechanically
isolated system. Units alone cannot rule it out, so Cue B is the most diagnostic type. \textbf{Cue
C} states the same quantity in equivalent units (200 W, 0.2 kW, 200 J/s, 200000 mW) and is admitted
only when the conversion is exact. Families are generated template-first. The symbolic problem and
its certificate exist before any text. Language models may paraphrase and vary variable names. They
never write physics or answers (Appendix~\ref{app:protocol}). Table~\ref{tab:bench} gives the
composition.

\begin{table}[t]
\caption{\bench{} composition and the sensitivity of Qwen2.5-7B-Instruct at the pre-registered
0.5-nat threshold. \emph{Families}: certified families. \emph{Labelled}: certified families included in the primary Qwen panel (excludes 15 families held
out to test transfer: 0 Cue A, 5 Cue B, 10 Cue C). \emph{Sensitive}: labelled families
whose log-probability drop reaches 0.5 nats. \emph{Mean drop}: mean $\Slp$ in nats over labelled
families.}
\label{tab:bench}
\small
\begin{tabularx}{\linewidth}{@{}Yd{1.6cm}d{1.6cm}d{2.2cm}d{1.9cm}@{}}
\toprule
\textbf{\underline{Cue type}} & \textbf{Families} & \textbf{Labelled} & \textbf{Sensitive (\%)} & \textbf{Mean drop} \\
\midrule
A: irrelevant variable & 35 & 35 & 12 (34.3) & 0.515 \\
B: unit-matched distractor & 105 & 100 & 53 (53.0) & 1.330 \\
C: equivalent units & 15 & 5 & 5 (100.0) & 9.022 \\
\midrule
\textbf{\underline{Domain}} & & & & \\
Mechanics & 91 & 86 & 46 (53.5) & 1.575 \\
Electrostatics and circuits & 43 & 37 & 17 (45.9) & 1.172 \\
Thermodynamics & 21 & 17 & 7 (41.2) & 1.014 \\
\midrule
\textbf{Total} & \textbf{155} & \textbf{140} & \textbf{70 (50.0)} & \textbf{1.401} \\
\bottomrule
\end{tabularx}
\end{table}

\subsection{Sensitivity label}\label{sec:label}

For a model with parameters $\theta$, let $p_{\theta}(y^{*} \mid x)$ be the probability it assigns
to the certified answer given prompt $x$, computed by teacher forcing, that is, by feeding the answer
tokens and reading off their probabilities without letting the model generate. The family's
sensitivity is the largest drop in log-probability relative to the base rendering,
\begin{equation}\label{eq:slp}
\Slp(\tau) = \max_{j \neq \mathrm{base}} \Bigl[ \log p_{\theta}\bigl(y^{*} \mid x_{\tau}(r, z_{\mathrm{base}}, a)\bigr) - \log p_{\theta}\bigl(y^{*} \mid x_{\tau}(r, z_j, a)\bigr) \Bigr],
\end{equation}
in nats, where the superscript lp stands for log-probability. For an ideal solver the four
log-probabilities are equal and $\Slp(\tau) = 0$, so $\Slp$ measures how far the model departs from the
invariance that Equation~\ref{eq:inv} certifies. We use $\Slp$ for three reasons.
\textbf{(a)} It is defined whenever the certified answer is known, independent of whether generated
text parses. \textbf{(b)} It is independent of decoding settings such as temperature and sampling
seed, because teacher forcing removes any role for sampling. \textbf{(c)} It is signed, so a family
whose confidence rises under the cue edit is not counted as sensitive. $\Slp$ is computed relative to a single, template-fixed
base rendering. A symmetric, base-free formulation is defined in Appendix~\ref{app:measures} but not
used for the primary label. A family is \emph{sensitive} when $\Slp \geq 0.5$. This
threshold was registered on 2026-06-14, before any probe was trained, from a 30-family pilot split.
Appendix~\ref{app:measures} defines two alternative statistics: the answer-flip rate, reported as
a secondary check, and the distribution-level divergence, not evaluated in this release.

\subsection{Hidden-state monitor}\label{sec:monitor-def}

The monitor reads the model before it answers. Let $h^{(\ell)}(x)$ be the residual-stream vector at
layer $\ell$ at the last token of prompt $x$. We average it over the family's renderings,
$\bar{h}_{\tau}^{(\ell)} = \tfrac{1}{4}\sum_{i=1}^{4} h^{(\ell)}(x_{\tau}(r,z_i,a))$. A
logistic-regression probe $f_{\phi}(\bar{h}_{\tau}^{(\ell)}) = \hat{s}$, with weights $\phi$, then
outputs the probability $\hat{s}$ that the family is sensitive. Every position read lies inside the
prompt, so the monitor is available before any token is generated. A single-rendering variant,
$f_{\phi}(h^{(\ell)}(x))$ without the family average, is evaluated separately in
Section~\ref{sec:monitor} and Appendix~\ref{app:full}.

\section{Experimental Setup}\label{sec:setup}

\paragraph{Models.} The primary model is \textbf{Qwen2.5-7B-Instruct} \citep{qwen2025qwen25technicalreport}
(28 layers, hidden width 3584). \textbf{Llama-3.1-8B-Instruct} \citep{grattafiori2024llama3herdmodels}
(32 layers) is the second architecture and \textbf{DeepSeek-R1-Distill-Qwen-32B}
\citep{deepseekai2026deepseekr1incentivizingreasoningcapability} the reasoning-trained comparison.
Activations are read and edited with TransformerLens \citep{nanda2022transformerlens} for the first
two and with forward hooks for DeepSeek. Extraction, patching and determinism were validated per
model before any experiment (Appendix~\ref{app:instr}).

\paragraph{Data.} Of the 155 certified families (620 prompts), 140 form the primary panel labelled
for Qwen. The other 15 were constructed after the probe was trained and are held out to test
transfer (Section~\ref{sec:monitor}) rather than included in the primary panel. The monitoring panel
holds 135 of the 140: all 70 sensitive families and 65 stable ones, after removing five stable
families whose Qwen outputs failed to parse.

\paragraph{Probe training.} The probe is $L_2$-regularised logistic regression
\citep{pedregosa2018scikitlearnmachinelearningpython}, swept over all layers. Regularisation strength
and standardisation are chosen by an inner cross-validation inside each training fold. Evaluation is
leave-one-family-out. The probe is trained on 134 families and scores the held-out one, 135 times.
All four renderings of a family, and any paraphrase or renamed copy of it, stay on the same side of
every split.

\paragraph{Comparators.} Eight predictors use no hidden states. \textbf{(a)} A \emph{fitted}
token-entropy proxy, the entropy of the model's next-token distribution passed through a classifier
trained on the same labels. \textbf{(b)} A combined model over nine raw measurements such as
entropy, answer log-probability, and prompt length, refit inside nested cross-validation.
\textbf{(c)} A black-box check that runs two renderings and compares outputs. \textbf{(d)} A
TF--IDF classifier on the prompt words. \textbf{(e)} Classifiers on the generated rationale and on
chain-of-thought text. \textbf{(f)} Self-consistency across sampled answers. \textbf{(g)} A
zero-shot LLM judge. A generic correctness probe, trained on hidden states to predict whether the
model answers correctly, is reported beside them.

\paragraph{Statistics.} Intervals come from a family bootstrap (10{,}000 resamples, seed 42) in
which whole families are resampled. Two predictors are compared by a \emph{paired} bootstrap that
scores both on the same resample in every iteration and reports the interval on their difference.

\paragraph{Causal intervention.} On a panel of 20 strongly sensitive Cue B families we knock out
chosen attention heads at prompt positions and recompute $\Slp$. The effect is the
\emph{normalised recovery}, the change in $\Slp$ divided by its original value, so 1.0 means the
drop is fully removed. Six controls run on the same families: \textbf{(a)} twenty random
perturbations of equal norm at the same site, \textbf{(b)} the same knockout at another layer,
\textbf{(c)} activations from an unrelated family, \textbf{(d)} from a family with the same answer,
\textbf{(e)} from a cue-bearing but stable family, and \textbf{(f)} accuracy on 40 held-out families
to detect damage.

\section{Results and Discussion}\label{sec:results}

\begin{figure}[t]
\centering
\includegraphics[width=0.40\linewidth]{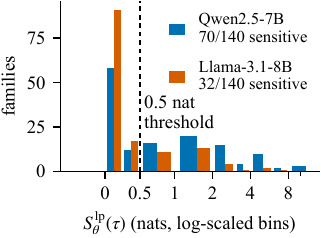}\hspace{0.02\linewidth}
\includegraphics[width=0.40\linewidth]{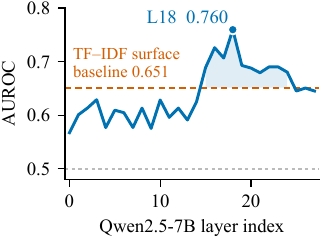}
\caption{\textbf{Left:} distribution of the log-probability drop $\Slp$ over the 140 labelled
families for both models on identical prompts. The dashed line is the pre-registered 0.5-nat
threshold, and Qwen's tail is much heavier. \textbf{Right:} probe AUROC at each layer of Qwen, with
families held out. The signal rises after layer 14, peaks at layer 18 (0.760), and the shaded area is
its margin over a classifier on the prompt's words (0.651).}
\label{fig:layer}
\end{figure}

\subsection{Sensitivity is common, model-specific and driven by numerical proximity}\label{sec:behaviour}

Qwen2.5-7B-Instruct is sensitive on 70 of 140 families (50.0\%, 95\% interval $[42.1, 58.6]$), with
mean $\Slp$ 1.401 nats and a maximum of 13.83 (Figure~\ref{fig:layer}, left). Llama-3.1-8B-Instruct
is sensitive on 32 of the same 140 families (22.9\%), and only 25 families are sensitive for both, so
sensitivity belongs to a pairing of model and problem. A probe trained to detect which cue value a
prompt contains scores 0.532, close to chance, which confirms that the label reflects the model's
behaviour. Cue B, where units cannot rule the distractor out,
has the highest rate (53.0\% against 34.3\% for Cue A), and the largest effects occur when the
distractor nearly matches the answer. The torque family of Appendix~\ref{app:examples} places a
6.9~N$\cdot$m distractor beside a 6.928~N$\cdot$m answer and gives 10.03 nats, and a frequency family
with 1.6 against 1.592~Hz gives 5.56. In a Cue A family an irrelevant 18~$^{\circ}$C wall temperature,
numerically equal to the 18~N answer, gives 2.30 nats. Across the full panel the linear correlation
between proximity and $\Slp$ is weak (Pearson 0.074 over 100 families, Appendix~\ref{app:additional}).
The effect concentrates in the near-match tail instead of forming a graded, broadly linear
relationship. The reasoning-trained DeepSeek model is
sensitive on 27.1\% of families and suppresses 36 of 53 Cue B positives while keeping all 5 Cue C
positives (Appendix~\ref{app:crossmodel}). This is consistent with the cue types engaging different
mechanisms, though DeepSeek's behaviour at this scoring position relative to its reasoning trace was
not independently confirmed (Appendix~\ref{app:instr}).

\subsection{Hidden states predict sensitivity before generation}\label{sec:monitor}

The layer-18 probe, averaged over a family's four renderings, reaches AUROC 0.760 (95\% interval
$[0.675, 0.838]$) with families held out (Figure~\ref{fig:layer}, right). Scored on a single
rendering instead, a separately swept probe reaches 0.747 at layer 16 (Appendix~\ref{app:full}).
The peak sits at 64\% of network depth in Qwen and at 69\% in
DeepSeek. The signal comes from processing inside the model. A TF--IDF classifier on the prompt
reaches 0.651 and a probe on the input embeddings 0.567. It is stable across labelling choices, with
AUROC between 0.722 and 0.782 as the threshold sweeps from 0.1 to 2.0 nats, and needs only part of
the family. A separate ablation, on a smaller Stage 6 panel distinct from the primary 135-family
sweep, gives 0.735 averaging two renderings against 0.731 with all four. Sensitive families are often hard ones (the correctness
probe's score correlates with the label at $-0.346$), and after removing the part of the score that
correctness explains, the probe keeps 0.738 (95\% interval $[0.650, 0.819]$). The effect replicates
in DeepSeek, where a probe on the spread of hidden states across renderings reaches 0.703 against
that model's own labels. A layer-16 probe frozen after training scores \textbf{0.820} (AUPRC 0.907) on fifteen families built
afterwards, split 5 Cue B and 10 Cue C. The pooled figure overstates within-type discrimination, since
Cue C predictions cluster high and Cue B cluster low. AUROC is 0.667 on Cue B alone and 0.6875 on Cue
C alone, each too small to resolve (Section~\ref{sec:limitations}).

\subsection{Comparison with non-activation baselines}\label{sec:compare}

\begin{figure}[t]
\centering
\includegraphics[width=0.72\linewidth]{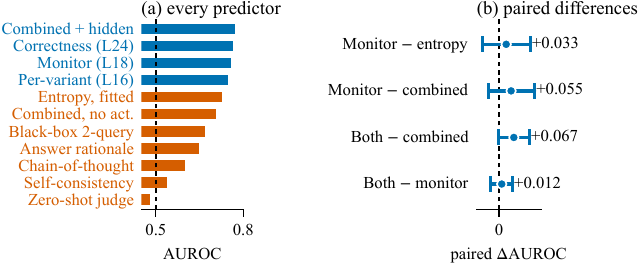}
\caption{\textbf{(a)} AUROC of every predictor on the 135-family panel. Blue predictors read hidden
states and orange ones use only the prompt or the model's outputs. \textbf{(b)} Paired differences
in AUROC with 95\% family-bootstrap intervals, both predictors scored on the same resampled families
in each of 10{,}000 iterations.}
\label{fig:incr}
\end{figure}

The probe exceeds every behavioural and text-based predictor (Figure~\ref{fig:incr}a), from the
black-box two-query check (0.668) through the chain-of-thought
classifier (0.599) to the zero-shot LLM judge (0.480). Two predictors come close. The fitted entropy
baseline reaches 0.727, and the paired difference of $+0.033$ has a 95\% interval of
$[-0.074, +0.141]$ (Figure~\ref{fig:incr}b), so on 135 families the two are statistically
indistinguishable. Descriptively, the probe ranks higher in 72\% of resamples. The correctness probe reaches 0.766,
consistent with the overlap between sensitivity and difficulty noted above. Adding hidden states to
the combined non-activation model raises its AUROC from 0.705 to 0.772, a gain of $+0.067$ whose
interval $[-0.001, +0.137]$ includes zero. The probe also reaches the lowest Brier score of any
single predictor, 0.204 against 0.214 for entropy and 0.229 for the combined model. Entropy
has the lower expected calibration error, however, and misses fewer sensitive families at a fixed
false-positive rate (0.643 against 0.714 at FPR 10\%, Appendix~\ref{app:full}).

These results give a clear practical reading. A fitted uncertainty signal is a strong and cheap
baseline, and a hidden-state probe matches it while adding calibrated probabilities before
generation. The two are natural companions in deployment. They also give a methodological one.
Probe studies commonly compare against random directions or surface features, and against those
baselines our probe shows a large margin (0.11 over TF--IDF, 0.16 over the chain-of-thought
classifier). The fitted, paired comparison is the one that tells a practitioner whether reading
activations is worth the access it requires, and we recommend it as the default.

\subsection{Causal intervention}\label{sec:causal}

\begin{figure}[t]
\centering
\includegraphics[width=0.60\linewidth]{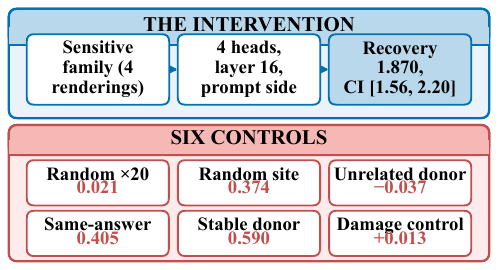}
\caption{The knockout intervention and its six controls on the Qwen2.5-7B-Instruct Cue B panel,
with the normalised recovery (or, for the damage control, the change in accuracy) each returned. The
knockout effect is roughly ninety times the random-direction control and three to five times the
random-site and donor controls.}
\label{fig:causaldesign}
\end{figure}

Knocking out heads 26, 24, 13 and 11 at layer 16 of Qwen gives a mean normalised recovery of
\textbf{1.870} (95\% interval $[1.560, 2.198]$) on the 20-family Cue B panel (Figure~\ref{fig:causaldesign}). A recovery above 1
means the intervention inverts the drop, so the model becomes more confident under the distractor
rendering than under its own base rendering. Capping each family's recovery at 1 gives 0.969. Head
11 carries almost all of the effect alone (1.848, leaving 0.022 for the other three heads combined),
and the window is one layer wide. The same head gives 0.540 at layer 15 and $-0.049$ at layer 17
(Appendix~\ref{app:causal}). The controls separate this effect from generic disruption: twenty
equal-norm random perturbations give 0.021 and are exceeded on 20 of 20 families, the random-site
control reaches 0.374, and the unrelated-family donor gives $-0.037$. Accuracy on 40 held-out
families moves from 0.700 to 0.713, inside a 5\% tolerance fixed in advance, showing the intervention
does not simply damage the model. The stable cue-bearing
donor recovers 0.590, 32\% of head 11's own 1.848, which attributes 68\% (0.681) of the effect
specifically to sensitivity. The same-answer donor, a complementary check on whether the effect
tracks answer identity, recovers even less (0.405). The same heads give 2.177 on ten held-out Cue A
and Cue C families (capped mean 0.702), a result carried almost entirely by Cue A. Four of the five
Cue C families recover below 0.35 (Appendix~\ref{app:causal}). The components the intervention
touches transfer reliably to Cue A, while the directions a linear probe reads are specific to each
cue type (cosine 0.018 between Cue A and Cue B). Adding the probe direction to stable families raises
their $\Slp$ steadily, but only 37\% cross the sensitivity threshold at twice the probe's own scale.
The heads, in other words, are necessary for the behaviour. The probe direction alone is only partly
sufficient for it. In Llama and DeepSeek the same recipe gives no reliable
effect at the sites tested (Appendix~\ref{app:causal}).

\subsection{Discussion}\label{sec:discussion}

\paragraph{Certification makes sensitivity a property of the model.} Because every family carries a
proof that its cue cannot change the answer, a confidence shift on \bench{} is attributable to the
model. The numbers then separate by model (Section~\ref{sec:behaviour}), with 50\% sensitive families
for Qwen and 23\% for Llama, only 25 shared, and a probe for the cue value itself scoring 0.532, so
the label carries no surface signature of the edit. The largest effects come from numerical
proximity, a distractor that nearly matches the answer (Section~\ref{sec:behaviour}). The cue types also differ in how
Qwen and DeepSeek respond to them (Appendix~\ref{app:crossmodel}), so a single robustness score
would hide structure that the certified, per-cue view exposes.

\paragraph{What the hidden state carries.} Since the probe performs on par with the fitted entropy
baseline at this sample size (Section~\ref{sec:compare}), the layer-18 state at the last prompt token
already holds much of what the model's next-token distribution will later express. The panel is not
powered to confirm this precisely, however (Section~\ref{sec:limitations}). The probe reads
internal residual-stream activations, while the entropy baseline reads only the output-token logits
from the same forward pass. The probe also reaches the lowest Brier score of any single predictor and
adds $+0.067$ to the combined non-activation model. Against the surface and embedding baselines common in probing
work (0.651 and 0.567), the probe's margin would read as 0.11 and 0.19 in place of 0.03. Sensitivity
overlaps with difficulty (correlation $-0.346$ with the correctness probe's score) and survives its
removal at 0.738, so the probe reads a property of the family beyond whether the model will answer
correctly.

\paragraph{A localised mechanism with cue-specific readouts.} The layer-16 heads
(Section~\ref{sec:causal}) generalise to Cue A but not reliably to Cue C, while the probe directions
for Cue A and Cue B are nearly orthogonal. The picture is one of components shared between Cue A and
Cue B that route the distractor's influence upstream, with cue-specific directions handling the
downstream readout. The two sufficiency
tests (Appendix~\ref{app:additional}) sharpen it, showing the heads are necessary for the behaviour
while the probe direction alone accounts for only part of it. The behaviour is therefore carried by
more of the activation state than one linear direction.
The depth at which the probe peaks (Section~\ref{sec:monitor}) does not by itself locate a causal
circuit, since the DeepSeek knockout at the matching depth recovers only a small fraction of the
effect (Appendix~\ref{app:causal}).

\paragraph{Where the monitor fails, and what to build next.} The probe's most confident errors are
families that instantiate the template of a sensitive problem without a near-matching number, and its
largest misses are strong positives whose signature is spread beyond the layer-18 mean feature
(Appendix~\ref{app:failures}). The probe therefore reads template class well and numerical proximity
poorly, which points to a numerically aware, per-rendering representation as the next monitor. The
flag it produces should be trusted to the extent that it beats a fitted uncertainty baseline under a
paired test.

\section{Conclusion}\label{sec:conclusion}

\bench{} makes sensitivity to irrelevant physics details a certified quantity: every family carries
a proof that its cue cannot change the answer, so a change in the model's confidence is attributable
to the model. On this benchmark, sensitivity is common, depends strongly on the model, peaks at
numerical near-matches, and is predictable from the model's own hidden state before it generates a
token, with a causal handle localised, in Qwen, to four attention heads at layer 16. For a practitioner building on this
result, the finding to carry forward is not the headline AUROC but the evaluation standard behind
it. A hidden-state monitor should be judged against the cheapest baseline that could plausibly
explain its accuracy, scored on the same resamples, not against a surface or random-direction
baseline that flatters it.

\section{Limitations}\label{sec:limitations}

\paragraph{Sample size.} The monitoring panel has 135 families with 70 positives. This resolves the
probe's margins of 0.11 to 0.28 over surface, text and judge baselines, and is too small to resolve
differences of 0.03 to 0.05 against fitted entropy. A larger panel would settle that comparison. The
reported layer-18 AUROC is the best of a 28-layer sweep, and held-out intervals are not adjusted for
that selection.

\paragraph{Scope of the causal result.} The head-knockout effect is established in Qwen2.5-7B-Instruct.
Normalised recovery divides by the original drop and becomes unstable on families with small drops.
A panel restricted to large drops in each model is needed to assess Llama and DeepSeek.

\paragraph{Coverage.} Cue C has five labelled families, so its large effects are illustrative, and
the same small-sample caveat limits the frozen-probe transfer check (Section~\ref{sec:monitor}).
Held-out families share a template cluster with training families in 71\% of cases (Appendix~\ref{app:integrity}), so leave-one-family-out is not leave-one-cluster-out.
Certification is symbolic and complete. A planned human validation will check whether each cue also
reads as irrelevant to a physicist. Holding out whole domains gives AUROC 0.750, statistically
indistinguishable from a permutation-shuffled null ($p = 0.368$) at this sample size, so
generalisation to unseen domains is not established (full scope in Appendix~\ref{app:scope}).

\subsection*{AI use statement}

An LLM-based coding assistant was used in preparing this paper. \textbf{Drafting}: it produced
initial drafts of prose sections from committed evidence artefacts, including interpreting what those
results show, under author direction on framing and scope. \textbf{Results}: it organised committed
numerical artefacts into tables and text. Every number was read from or recomputed from a committed
artefact. \textbf{Literature and references}: it proposed related work and verified each entry
against two independent sources before retrieving BibTeX programmatically. No entry was written from
model memory. \textbf{Editing and code}: it suggested structure and wording and assisted with
analysis and verification code. \textbf{Benchmark data}: within the benchmark pipeline, language
models contributed to synthetic data generation by paraphrasing renderings, varying variable names
and proposing candidate failure cases for human review, and could not write physics, determine ground
truth or define correct answers. The assistant did not design the study, choose the sensitivity
measure, set the threshold, or run the experiments. The authors take full responsibility for all
content.

\subsection*{Ethics statement}

This work evaluates published open-weight models on synthetic physics problems generated from
symbolic templates. It involves no human-subjects data, no personal data and no scraped web content.
The planned human validation will compensate annotators and supply the qualification and calibration
materials already committed. The methods are diagnostic and do not enhance model capabilities.

\subsection*{Reproducibility statement}

Every reported number derives from committed artefacts. The release includes the benchmark manifest
with cue type, domain and certificate for each family; per-family labels for both primary models;
probe predictions; all baseline scores; causal patching results; and all control summaries.
Table~\ref{tab:bench} recomputes exactly from the released per-family table. Bootstraps use seed 42
with 10{,}000 resamples and the family as the resampling unit. The threshold, its rationale, and its
registration date and commit are recorded in the construct specification. Model identities, layer
counts, hidden widths and hook back-ends are in a committed model registry, and the environment is
pinned by a lock file. Teacher-forced log-probabilities are scored on each model's native chat
template, with an answer-format instruction that produces the \texttt{Answer: [value] [unit]}
pattern seen in the released generations (Appendix~\ref{app:examples}). The appendices give full
methodology, a complete worked family, and every result table.

\bibliography{refs}
\bibliographystyle{iclr2027_conference}

\appendix
\section{A Family End to End: Files and Worked Example}\label{app:examples}

Section~\ref{sec:bench} describes what a model sees in general. This appendix reproduces the concrete
files for family \texttt{CM\_B\_UM\_059}, the records that score it, exactly as they are stored. It
then works the sensitivity computation by hand so that every number in the paper's
headline example can be recomputed from what is printed here.

\paragraph{File 1: the canonical template.} Every family begins as a YAML template written before any
prompt exists. The \texttt{governing\_equation\_sympy} field is the physics. \texttt{parameters} are
the relevant variables $r$ with their values and units. \texttt{correct\_answer} is $y^{*}$ with the
solver's own numerical check. \texttt{cue\_slot} declares the cue $z$: the symbol the verifier must
prove absent (\texttt{distractor\_symbol}), the four values it takes, and which of them is the base
rendering. \texttt{prompt\_template} is the rendering with the cue in braces, and
\texttt{cue\_sentence} records the span that is allowed to differ across the family.

\begin{lstlisting}
template_id: CM_B_UM_059
domain: mechanics
cue_type: nongoverning_distractor
sequence: 59
created_date: '2026-06-15'
created_by: agent
governing_law: torque_magnitude
governing_equation_latex: \tau_A = r_A F_A \sin\theta
governing_equation_sympy: r_A * F_A * sin(theta_deg * pi / 180)
target_quantity: torque_on_system_A
target_units: N*m
parameters:
  r_A:
    value: 0.4
    units: m
    description: lever arm length
  F_A:
    value: 20.0
    units: N
    description: applied force
  theta_deg:
    value: 60.0
    units: deg
    description: force angle
correct_answer:
  value: 6.928203230275509
  units: N*m
  display: 6.928 N*m
  sympy_check: abs((r_A * F_A * sin(theta_deg * pi / 180)) - 6.928203230275509) <
    6.9282032302755086e-06
cue_slot:
  name: tau_B_display
  description: Torque of separate system B, unit-matched with the target torque
  type: numerical
  irrelevance_proof: null
  distractor_symbol: tau_B
  distractor_units: N*m
  nongoverning_proof: The governing expression "r_A * F_A * sin(theta_deg * pi / 180)"
    has free symbols {r_A, F_A, theta_deg}; the distractor symbol tau_B is absent
    from the governing equation. System B is mechanically isolated from system A.
  values:
  - id: 0
    value: '3.5'
    display: '3.5'
  - id: 1
    value: '6.9'
    display: '6.9'
  - id: 2
    value: '10.0'
    display: '10.0'
  - id: 3
    value: '14.0'
    display: '14.0'
  base_variant_id: 0
prompt_template:
  context: System A has a lever arm of {r_A} m with a force of {F_A} N applied at
    60 degrees to the lever arm. In an adjacent workshop, a separate wrench system
    B produces a torque of {tau_B_display} N.m, but it is mechanically isolated from
    system A.
  question: What torque acts on system A?
  full_template: 'System A has a lever arm of {r_A} m with a force of {F_A} N applied
    at 60 degrees to the lever arm. In an adjacent workshop, a separate wrench system
    B produces a torque of {tau_B_display} N.m, but it is mechanically isolated from
    system A.


    What torque acts on system A?'
  cue_span_is_whole_sentence: false
  cue_sentence: In an adjacent workshop, a separate wrench system B produces a torque
    of {tau_B_display} N.m, but it is mechanically isolated from system A.
verifier:
  method: sympy_numeric
  sympy_check_code: 'r_A = 0.4

    F_A = 20.0

    theta_deg = 60.0

    result = r_A * F_A * sin(theta_deg * pi / 180)

    assert abs(result - 6.928203230275509) < 6.9282032302755086e-06, f"Answer check
    failed: {result}"'
  cue_independence_method: sympy_free_symbols
validation:
  verifier_certified: true
  pi_validated: true
  v2_validated: false
  kappa_contribution: null
  exclusion_flags: []
  notes: ''
topic_tags:
- mechanics
- torque
- rotation
- unit_matched_cue_b
- near_match
difficulty: introductory
notes: ''
\end{lstlisting}

\paragraph{File 2: the verifier's output.} The verifier evaluates the four checks of
Section~\ref{sec:bench} on the template and writes one record. The \texttt{cue\_independence} check is
Equation~\ref{eq:cert}: the free-symbol set of the solved expression is
$\{\texttt{F\_A}, \texttt{r\_A}, \texttt{theta\_deg}\}$ and \texttt{tau\_B} is not in it. A family
enters the benchmark only when \texttt{all\_passed} is true.

\begin{lstlisting}
{
  "all_passed": true,
  "checks": [
    {
      "check_name": "answer_correctness",
      "detail": "evaluated=6.92820323027551, expected=6.928203230275509, tolerance=6.9282032302755086e-06",
      "passed": true
    },
    {
      "check_name": "cue_independence",
      "detail": "cue_slot.distractor_symbol absent from governing-equation free symbols ['F_A', 'r_A', 'theta_deg']",
      "passed": true
    },
    {
      "check_name": "physical_plausibility",
      "detail": "all parameter values are finite and plausible",
      "passed": true
    },
    {
      "check_name": "correct_answer_parseable",
      "detail": "Parsed canonical answer as '6.928 N*m'.",
      "passed": true
    }
  ],
  "template_id": "CM_B_UM_059"
}
\end{lstlisting}

\paragraph{File 3: the rendered family.} These are the four prompts a model receives, each carrying
its cue value and the shared certified answer. This is the file the behavioural run reads. It holds no physics, only
text, so the label it produces is a property of the model.

\begin{lstlisting}
{
  "correct_answer": "6.928 N*m",
  "cue_type": "nongoverning_distractor",
  "domain": "mechanics",
  "template_id": "CM_B_UM_059",
  "variants": [
    {
      "correct_answer": "6.928 N*m",
      "cue_sentence": "In an adjacent workshop, a separate wrench system B produces a torque of 3.5 N.m, but it is mechanically isolated from system A.",
      "cue_value": "3.5",
      "prompt": "System A has a lever arm of 0.4 m with a force of 20.0 N applied at 60 degrees to the lever arm. In an adjacent workshop, a separate wrench system B produces a torque of 3.5 N.m, but it is mechanically isolated from system A.\n\nWhat torque acts on system A?",
      "variant_id": 0
    },
    {
      "correct_answer": "6.928 N*m",
      "cue_sentence": "In an adjacent workshop, a separate wrench system B produces a torque of 6.9 N.m, but it is mechanically isolated from system A.",
      "cue_value": "6.9",
      "prompt": "System A has a lever arm of 0.4 m with a force of 20.0 N applied at 60 degrees to the lever arm. In an adjacent workshop, a separate wrench system B produces a torque of 6.9 N.m, but it is mechanically isolated from system A.\n\nWhat torque acts on system A?",
      "variant_id": 1
    },
    {
      "correct_answer": "6.928 N*m",
      "cue_sentence": "In an adjacent workshop, a separate wrench system B produces a torque of 10.0 N.m, but it is mechanically isolated from system A.",
      "cue_value": "10.0",
      "prompt": "System A has a lever arm of 0.4 m with a force of 20.0 N applied at 60 degrees to the lever arm. In an adjacent workshop, a separate wrench system B produces a torque of 10.0 N.m, but it is mechanically isolated from system A.\n\nWhat torque acts on system A?",
      "variant_id": 2
    },
    {
      "correct_answer": "6.928 N*m",
      "cue_sentence": "In an adjacent workshop, a separate wrench system B produces a torque of 14.0 N.m, but it is mechanically isolated from system A.",
      "cue_value": "14.0",
      "prompt": "System A has a lever arm of 0.4 m with a force of 20.0 N applied at 60 degrees to the lever arm. In an adjacent workshop, a separate wrench system B produces a torque of 14.0 N.m, but it is mechanically isolated from system A.\n\nWhat torque acts on system A?",
      "variant_id": 3
    }
  ],
  "verifier_certified": true
}
\end{lstlisting}

\paragraph{File 4: one behavioural record.} The behavioural run writes one record per model per
rendering. The record for rendering 1 under Qwen2.5-7B-Instruct is reproduced in full. The fields that
matter for the label are \texttt{logprob\_correct\_answer}, the teacher-forced log-probability of the
certified answer, and \texttt{parsed\_answer\_canonical}, the parser's reading of what the model
generated.

\begin{lstlisting}
{
  "answer_correct_postanalysis": null,
  "correct_answer": "6.928 N*m",
  "cue_type": "nongoverning_distractor",
  "cue_value": "6.9",
  "domain": "mechanics",
  "dry_run": false,
  "generated_text": "Answer: 6.84 N.m",
  "git_commit": "ffeeeff",
  "logprob_correct_answer": -55.5625,
  "model_name": "Qwen/Qwen2.5-7B-Instruct",
  "model_role": "PRIMARY_DENSE",
  "parse_confidence": 0.98,
  "parse_confident": true,
  "parsed_answer": "6.84 N*m",
  "parsed_answer_canonical": "6.84 N*m",
  "prompt": "System A has a lever arm of 0.4 m with a force of 20.0 N applied at 60 degrees to the lever arm. In an adjacent workshop, a separate wrench system B produces a torque of 6.9 N.m, but it is mechanically isolated from system A.\n\nWhat torque acts on system A?",
  "record_type": "variant_record",
  "script_name": "run_behavioural.py",
  "seed": 42,
  "stage": 4,
  "template_id": "CM_B_UM_059",
  "variant_id": 1
}
\end{lstlisting}

\paragraph{Worked computation.} Table~\ref{tab:worked} collects the four records for each primary
model. The log-probability drop of Equation~\ref{eq:slp} takes the base rendering (rendering 0, cue
value 3.5) as reference and reports the largest drop among the other three:
\begin{align*}
\text{Qwen:}\quad & 10.031,\; -0.531,\; 0.938 \;\text{ for renderings } 1, 2, 3
\;\;\Rightarrow\;\; \Slp = 10.031 \;\text{ (at rendering 1)}, \\
\text{Llama:}\quad & 2.000,\; -0.781,\; 0.188 \;\text{ for renderings } 1, 2, 3
\;\;\Rightarrow\;\; \Slp = 2.000 \;\text{ (at rendering 1)},
\end{align*}
each drop being the base log-probability minus the rendering's, for example
$(-45.531) - (-55.562) = 10.031$ for Qwen at rendering 1. Drops are computed from unrounded raw
log-probabilities, so subtracting the rounded values printed in Table~\ref{tab:worked} can give a
result that differs from the reported drop by up to 0.001 nats.
Both exceed the pre-registered 0.5-nat threshold, so the family is labelled sensitive for both
models. The values match the released per-family table (\texttt{qwen\_S\_lp} 10.03125,
\texttt{llama\_S\_lp} 2.0). The answer-flip statistic of Equation~\ref{eq:flip} counts pairs of
renderings whose parsed answers differ. Qwen's four answers (13.86, 6.84, 8, 8) differ on five of the
six pairs, giving $\Sflip = 5/6 = 0.833$. Llama's four identical answers give $\Sflip = 0$. Both
match the table (\texttt{qwen\_hat\_S} 0.8333, \texttt{llama\_hat\_S} 0.0). The largest drop for both
models falls at rendering 1, where the distractor 6.9 lies 0.4\% from the certified answer 6.928, which
is the near-match mechanism of Section~\ref{sec:behaviour}. Qwen is wrong on every rendering and
Llama is wrong on every rendering, so this family is sensitive without being correct, which is
exactly the case the certified-invariance design exists to separate from accuracy loss.

\begin{table}[h]
\caption{The four behavioural records for \texttt{CM\_B\_UM\_059}, both primary models. Log-probability
is of the certified answer ``6.928 N*m'' under teacher forcing. The generated text is what the model
produced when allowed to answer freely.}
\label{tab:worked}
\small
\begin{tabularx}{\linewidth}{@{}d{1.2cm}d{0.9cm}Yd{1.9cm}Yd{1.9cm}@{}}
\toprule
\textbf{Rendering} & \textbf{Cue} & \textbf{Qwen generated} & \textbf{Qwen log\,$p$} & \textbf{Llama generated} & \textbf{Llama log\,$p$} \\
\midrule
0 (base) & 3.5 & Answer: 13.86 N.m & $-45.531$ & Answer: 8 N.m. & $-32.938$ \\
1 & 6.9 & Answer: 6.84 N.m & $-55.562$ & Answer: 8 N.m. & $-34.938$ \\
2 & 10.0 & Answer: 8.0 N.m & $-45.000$ & Answer: 8 N.m. & $-32.156$ \\
3 & 14.0 & Answer: 8.0 N.m & $-46.469$ & Answer: 8.0 N.m. & $-33.125$ \\
\bottomrule
\end{tabularx}
\end{table}

\paragraph{The other two cue types.} \texttt{CM\_A\_STD\_005} (Cue A) asks for a centripetal
force of $18.0$ N from $m = 2.0$ kg, $v = 3.0$ m/s, $r = 1.0$ m, with governing expression
\texttt{m * v**2 / r}. It varies an irrelevant outer-track-wall temperature over 9, 18, 27 and 45
$^{\circ}$C. Its four prompts read \emph{Object A of mass 2.0 kg travels at 3.0 m/s around a circle
of radius 1.0 m. The outer track wall temperature is 9 $^{\circ}$C. What centripetal force magnitude
is required for object A?} with only the temperature changing. The irrelevance proof records that the
governing expression depends only on $m$, $v$ and $r$. Rendering 1 sets the temperature to 18,
numerically equal to the correct force in different units, and the family shows $\Slp = 2.30$ nats,
the drop falling at exactly that rendering. Qwen generates the wrong answer (6.0 N) on every rendering, so
the effect is a shift in the probability it assigns to the correct answer, with no answer flip.
It is nonetheless the cleanest evidence that a numeral, not a physical quantity, is what the model
responds to. \texttt{CM\_C\_004} (Cue C) reads \emph{Meter A reports the power as 200 W. What is that
power in watts?} with the power rendered as \texttt{200 W}, \texttt{0.2 kW}, \texttt{200 J/s} or
\texttt{200000 mW}. The governing expression is the identity $P = P_{\mathrm{base}}$, so invariance
is exact by construction. Its sensitivity is large ($\Slp = 13.55$ for Qwen, with the largest drop at
the \texttt{0.2 kW} rendering) because the model is divided between emitting the canonical value and
the prompt-displayed one. This is a Level-2 instability of a different kind from Cue B distractor
confusion, which is why per-cue results are reported separately throughout.

\section{Alternative Sensitivity Measures}\label{app:measures}

Section~\ref{sec:label} uses the log-probability drop $\Slp$ as the label. Two other statistics
describe the same behaviour and serve as secondary checks.

\paragraph{Answer-flip rate.} The fraction of rendering pairs whose parsed answers differ,
\begin{equation}\label{eq:flip}
\Sflip(\tau) = \frac{1}{\binom{m}{2}} \sum_{i<j} \mathbf{1}\bigl[\, \hat{y}_{\theta}(x_{\tau}(r,z_i,a)) \neq \hat{y}_{\theta}(x_{\tau}(r,z_j,a)) \,\bigr],
\end{equation}
where $m = 4$ is the number of renderings, $\hat{y}_{\theta}(x)$ is the model's generated answer to
prompt $x$ after parsing, $\mathbf{1}[\cdot]$ is 1 when its condition holds and 0 otherwise, and
$\binom{m}{2} = 6$ is the number of rendering pairs. Pairs with an unparseable answer are dropped and
the denominator reduced accordingly. $\Sflip$ depends on the parser and on decoding, and it ignores
confidence shifts that leave the top answer unchanged, as in the kinetic-energy family of the
introduction. These are the reasons it is secondary.

\paragraph{Distribution-level divergence.} The mean pairwise Jensen--Shannon divergence between
answer distributions over a normalised answer vocabulary,
\begin{equation}\label{eq:jsd}
\Sjsd(\tau) = \frac{2}{m(m-1)} \sum_{i<j} D_{\mathrm{JS}}\!\Bigl( p_{\theta}\bigl(y \mid x_{\tau}(r,z_i,a)\bigr),\; p_{\theta}\bigl(y \mid x_{\tau}(r,z_j,a)\bigr) \Bigr),
\end{equation}
where $p_{\theta}(y \mid x)$ is the probability of answer string $y$ given prompt $x$ and
$D_{\mathrm{JS}}$ is a symmetric measure, bounded between 0 and $\ln 2$ nats, of how far apart two
distributions are.
Estimating $p_{\theta}$ over open-ended generation needs top-$k$ token probabilities restricted to an
answer vocabulary or repeated sampling, both of which add cost and an approximation error that the
certificate does not bound. $\Slp$ avoids both by scoring the one certified answer directly.

\section{Generation Protocol}\label{app:protocol}

\paragraph{Eight steps.} Section~\ref{sec:bench} states the principle. Every family is generated
under the same eight-step protocol. \textbf{(a)} Define a canonical symbolic problem: the relevant variables $r$, assumptions $a$,
governing principle $g$, and the solver-derived answer $y^{*}$. \textbf{(b)} Designate the cue $z$
and write the symbolic proof or executable certificate that $y^{*}$ is invariant to it.
\textbf{(c)} Instantiate numerical settings for $r$, varying $z$ across four values per
family. \textbf{(d)} Render matched natural-language prompts such that only the recorded cue span
changes across family members. \textbf{(e)} Generate paraphrased and variable-renamed variants for
robustness testing. \textbf{(f)} Run automatic checks for dimensional consistency, answer invariance
through the parser, assumption completeness, and surface-statistic matching on length, number count,
variable count and unit-token count. \textbf{(g)} Run adversarial artefact analysis, prompting a
separate model to predict the sensitivity class from the prompt alone and inspecting the features it
uses. \textbf{(h)} Specify dual human validation as the final protocol step. The pilot has not yet
been run for the families in this release (Appendix~\ref{app:integrity}). Language models are \emph{permitted}
to paraphrase, diversify variable names, discover adversarial artefacts, and propose candidate
failure cases for human review. They are \emph{prohibited} from inventing physics, determining ground
truth, defining the correct answer, or generating problem semantics that are then adopted without
verification. The cue span is stored explicitly with each family so that artefact checks can confirm
members differ only within it, and the construction-stage artefact report records 30 of 30 templates passing
cue-span isolation.

\section{Instrumentation Validation}\label{app:instr}

Section~\ref{sec:setup} describes the pass/fail gates run before any experiment.
Table~\ref{tab:instr} is the validation record. TransformerLens serves Qwen and Llama. DeepSeek is read through direct
forward hooks, functions attached to a layer to read or overwrite its output. The DeepSeek entry carries one caveat
recorded at the time and never resolved: the validation prompt did not emit \texttt{<think>} tags.
As a result, the reasoning-trace assumption was never confirmed on that model, though no result in
this paper depends on it. Forward-pass cost was 0.343 s for Qwen and 0.403 s
for Llama at the benchmark's prompt lengths. The exhaustive activation-storage upper bound across
both primary models, all layers and both prompt positions was estimated at 1.57 TB in float16. This
is why extraction is staged by site.

\begin{table}[h]
\caption{Instrumentation validation. All checks are pass/fail gates run before experiments.}
\label{tab:instr}
\small
\begin{tabularx}{\linewidth}{@{}Yd{1.4cm}d{1.4cm}d{2.2cm}d{2.4cm}Y@{}}
\toprule
\textbf{Model} & \textbf{Layers} & \textbf{$d_{\mathrm{model}}$} & \textbf{Backend} & \textbf{Patch $\Delta$logit} & \textbf{Status} \\
\midrule
Qwen2.5-7B-Instruct & 28 & 3584 & TransformerLens & 30.55 & extraction, patching, determinism pass \\
Llama-3.1-8B-Instruct & 32 & 4096 & TransformerLens & 25.72 & extraction, patching, determinism pass \\
DeepSeek-R1-Distill-Qwen-32B & 64 & 5120 & direct hooks & n/a & log-probability and determinism pass \\
\bottomrule
\end{tabularx}
\end{table}

\section{Full Monitoring Tables}\label{app:full}

\begin{figure}[h]
\centering
\includegraphics[width=\linewidth]{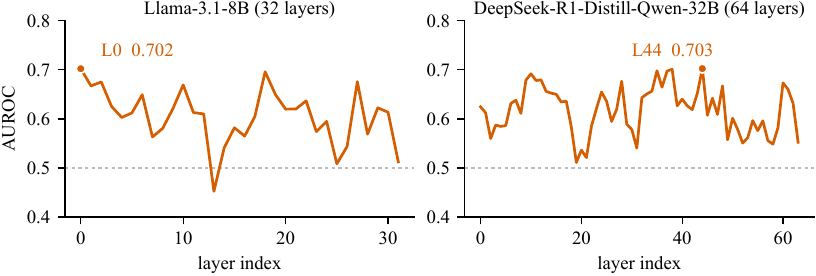}
\caption{Variance-probe AUROC by layer for the two other models, leave-one-family-out against each
model's own behaviour. DeepSeek peaks at layer 44 of 64 (0.703). Llama's best layer is layer 0, the
token embeddings before any processing (0.702), a lexical signal distinct from a late-layer
representation, which is why no Llama monitoring claim is made in the body.}
\label{fig:app-crossmodel}
\end{figure}

Table~\ref{tab:probes} lists every probe configuration evaluated, and Table~\ref{tab:baselines}
every non-activation and textual comparator, on the 135-family panel of Sections~\ref{sec:monitor}
and~\ref{sec:compare}. Figure~\ref{fig:app-crossmodel} plots the corresponding layer sweep for Llama
and DeepSeek. The DeepSeek result is the corresponding row of Table~\ref{tab:probes}. Llama's peak,
a lexical layer-0 signal, is not tabulated as a formal probe configuration.

\begin{figure}[h]
\centering
\includegraphics[width=0.485\linewidth]{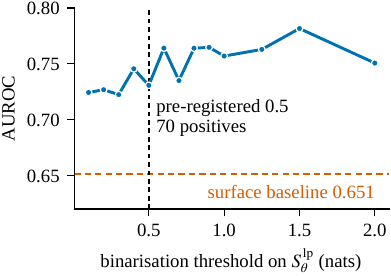}\hfill
\includegraphics[width=0.485\linewidth]{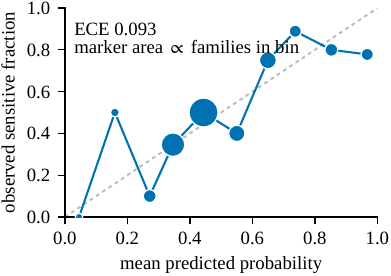}
\caption{\textbf{Left:} monitor AUROC against the binarisation threshold applied to $\Slp$, from the
Stage 9 sweep. The pre-registered 0.5 is marked, and the sweep's maximum (0.782 at 1.5) lies above it.
\textbf{Right:} reliability diagram of the layer-18 monitor in ten probability bins (expected
calibration error, ECE, 0.093 in this run). Marker area is proportional to the number of families in the bin, and the diagonal is
perfect calibration.}
\label{fig:app-threshold}
\end{figure}

\paragraph{Threshold sensitivity.} In the sweep's own run, AUROC is 0.724 at a threshold of 0.1
(87 positives), 0.727 at 0.2, 0.722 at 0.3, 0.746 at 0.4 and 0.731 at the pre-registered 0.5
(70 positives). This is against a maximum of 0.782 across the sweep and a fixed surface reference of
0.651 (Figure~\ref{fig:app-threshold}, left). The pre-registered threshold is not the most favourable
choice available. Figure~\ref{fig:app-threshold} (right) gives the reliability diagram for the
layer-18 monitor at that threshold.

\begin{table}[h]
\caption{All probe configurations evaluated, leave-one-family-out unless noted. Site abbreviations:
RP-LP is residual stream at the last prompt token, RP-CT is residual stream at the cue token.}
\label{tab:probes}
\small
\begin{tabularx}{\linewidth}{@{}Yd{1.5cm}d{1.3cm}d{1.4cm}d{3.4cm}@{}}
\toprule
\textbf{Configuration} & \textbf{AUROC} & \textbf{Layer} & \textbf{Site} & \textbf{95\% CI} \\
\midrule
Mean probe, full sweep (primary) & 0.7596 & 18 & RP-LP & $[0.6733, 0.8405]$ \\
Variance-ensemble probe (PCA-50, L15--20) & 0.7475 & 18 & RP-LP & $[0.6657, 0.8290]$ \\
Per-variant probe sweep & 0.7465 & 16 & RP-LP & $[0.7035, 0.7871]$ \\
MLP probe & 0.7424 & 18 & RP-LP & $[0.6546, 0.8207]$ \\
Correctness-residualised monitor & 0.7376 & 18 & RP-LP & $[0.6504, 0.8185]$ \\
Domain-generalisation probe & 0.7343 & 18 & RP-LP & $[0.6453, 0.8151]$ \\
Variance probe (Stage 6 panel) & 0.7308 & 18 & RP-LP & $[0.6424, 0.8086]$ \\
Cue-B-only probe & 0.7284 & 22 & RP-LP & $[0.6190, 0.8217]$ \\
DeepSeek variance probe & 0.7025 & 44 & RP-LP & $[0.6024, 0.7974]$ \\
Head-output probe (head 15) & 0.7040 & 10 & attn out & not computed \\
Head-output probe (head 14) & 0.7051 & 24 & attn out & not computed \\
Multi-site probe (7168-dim, 4 sites) & 0.6876 & 16 & multi & not computed \\
Mean probe at cue-token site & 0.6167 & 26 & RP-CT & $[0.5184, 0.7170]$ \\
Embedding-layer probe & 0.5673 & 0 & RP-LP & not computed \\
Prompt-condition probe (Level-1 control) & 0.5324 & 20 & RP-LP & not computed \\
\bottomrule
\end{tabularx}
\end{table}

\begin{table}[h]
\caption{All non-activation and textual comparators on the 135-family monitoring panel, with the
hidden-state correctness probe shown for reference.}
\label{tab:baselines}
\small
\begin{tabularx}{\linewidth}{@{}Yd{1.6cm}Y@{}}
\toprule
\textbf{Comparator} & \textbf{AUROC} & \textbf{Access required} \\
\midrule
Generic correctness probe (L24) & 0.7657 & hidden states \\
Token entropy proxy, fitted & 0.7270 & one forward pass \\
Combined non-activation (9 raw features) & 0.7048 & one forward pass \\
Black-box two-query counterfactual check & 0.6684 & two queries \\
Surface TF--IDF on prompt text & 0.6514 & prompt only \\
Answer-and-rationale TF--IDF classifier & 0.6464 & generated text \\
Chain-of-thought text classifier & 0.5990 & generated text \\
Self-consistency over sampled answers & 0.5369 & repeated sampling \\
Zero-shot LLM judge (mean over settings) & 0.4797 & judge model \\
\bottomrule
\end{tabularx}
\end{table}

\begin{table}[h]
\caption{Calibration and operating-point metrics on the 135-family panel. Brier score is the mean
squared error of the predicted probability. ECE (expected calibration error) is the average gap
between predicted probability and observed frequency over probability bins. FNR@FPR.10 is the
fraction of sensitive families missed when the threshold flags 10\% of stable families. Lower is
better for all three.}
\label{tab:incr}
\small
\begin{tabularx}{\linewidth}{@{}Yd{1.7cm}d{1.7cm}d{1.7cm}d{1.7cm}d{2.2cm}@{}}
\toprule
\textbf{Predictor} & \textbf{AUROC} & \textbf{AUPRC} & \textbf{Brier} & \textbf{ECE} & \textbf{FNR@FPR.10} \\
\midrule
Combined non-activation & 0.7048 & 0.6986 & 0.2288 & 0.1063 & 0.857 \\
Token entropy, fitted & 0.7270 & 0.7551 & 0.2140 & \textbf{0.0907} & \textbf{0.643} \\
Hidden-state monitor (L18) & 0.7596 & 0.7289 & \textbf{0.2036} & 0.1041 & 0.714 \\
Combined $+$ hidden state & \textbf{0.7719} & \textbf{0.7552} & 0.2244 & 0.1713 & 0.714 \\
\bottomrule
\end{tabularx}
\end{table}

\paragraph{Calibration and operating points.} Table~\ref{tab:incr} adds the operating-point view to
Section~\ref{sec:compare}. The hidden-state monitor has the lowest Brier score. Combining hidden states
with the non-activation features gives the highest AUROC and AUPRC and the highest ECE (0.171 against
0.091 for fitted entropy). At a fixed threshold flagging 10\% of stable families, the fitted entropy
baseline misses 64.3\% of sensitive families and the combination 71.4\%. For a deployment that acts at
a fixed threshold, the fitted entropy baseline and the monitor together are the practical choice.

\paragraph{Combined non-activation baseline.} The combined baseline of Section~\ref{sec:compare}
is fitted under nested stratified family-level cross-validation with 5 outer and 4 inner folds, seed
42, the canonical family as the resampling unit, and per-fold selected regularisation strengths of
$\{0.01, 0.01, 100.0, 10.0, 0.01\}$. Its nine features are the entropy proxy, mean correct-answer
log-probability, answer magnitude in $\log_{10}$, equation character count, equation symbol count,
parameter count, and prompt character, number and word counts. All of them are raw measurements
refit inside the cross-validation, which the artefact records explicitly.

\section{Causal Results in Full}\label{app:causal}

\begin{figure}[h]
\centering
\includegraphics[width=0.62\linewidth]{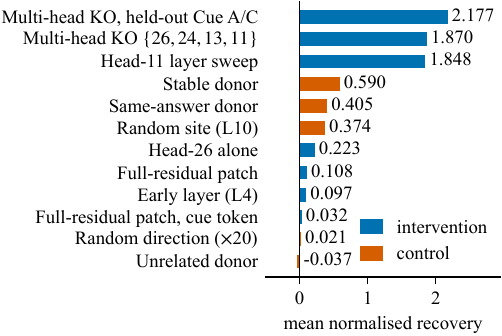}
\caption{Every intervention and control on the 20-family Cue B panel in Qwen2.5-7B. The knockout
effect exceeds the norm-matched random-direction control by roughly two orders of magnitude and
exceeds the random-site, stable-donor and same-answer-donor controls by a factor of three to five.
The unrelated-family donor is correctly negative.}
\label{fig:causal}
\end{figure}

\begin{figure}[h]
\centering
\includegraphics[width=\linewidth]{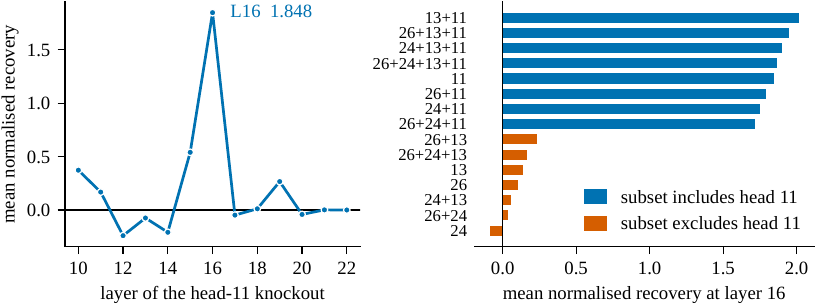}
\caption{\textbf{Left:} knocking out head 11 alone at each layer from 10 to 22 on the 20-family Cue B
panel. The causal window is one layer wide: 1.848 at layer 16 against 0.540 at layer 15 and $-0.049$
at layer 17. \textbf{Right:} every subset of the four heads knocked out at layer 16. Every subset
containing head 11 recovers between 1.71 and 2.01. No subset without it exceeds 0.24.}
\label{fig:app-h11}
\end{figure}

\begin{figure}[h]
\centering
\includegraphics[width=\linewidth]{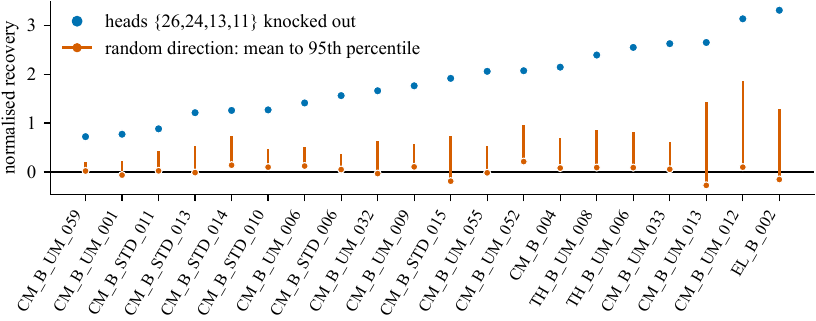}
\caption{The random-direction control family by family, sorted by the knockout effect. For each
family the actual four-head knockout recovery is compared with the mean and 95th percentile of twenty
norm-matched random perturbations at the same site. The knockout exceeds the 95th percentile on 20 of
20 families. Recovery is heavy-tailed because it divides by the family's original $\Slp$.}
\label{fig:app-rdc}
\end{figure}

Figure~\ref{fig:causal} plots and Table~\ref{tab:causal} tabulates every intervention and control of
Section~\ref{sec:causal}, together with the three interventions on other architectures.
Figure~\ref{fig:app-h11} isolates the layer window and head-subset structure behind the head-11
result, and Figure~\ref{fig:app-rdc} gives the random-direction control family by family.

\begin{table}[h]
\caption{All interventions and controls, ordered by mean normalised recovery of $\Slp$ on the
20-family Cue B panel unless noted. Recovery divides by the original $\Slp$ and is heavy-tailed when
that denominator is small.}
\label{tab:causal}
\small
\begin{tabularx}{\linewidth}{@{}Yd{2.4cm}d{1.8cm}Y@{}}
\toprule
\textbf{Intervention} & \textbf{Site} & \textbf{Recovery} & \textbf{Reading} \\
\midrule
\multicolumn{4}{@{}l}{\textit{Qwen2.5-7B-Instruct}} \\
Multi-head knockout, held-out Cue A/C panel & L16 & 2.177 & Generalises to Cue A only \\
Multi-head knockout $\{26,24,13,11\}$ & L16 & 1.870 & Primary result; clipped 0.969 \\
Head-11 layer sweep & L16 & 1.848 & Most of the effect is head 11 \\
Cue-present stable donor & L16 & 0.590 & Specificity ratio 3.13 \\
Same-answer donor & L16 & 0.405 & Specificity ratio 4.57 \\
Random-site control & L10 & 0.374 & Localisation control \\
Head-26 knockout alone & L16 & 0.223 & Single head insufficient \\
Full-residual patching & last prompt & 0.108 & Blunt instrument underperforms \\
Early-layer multi-head knockout & L4 & 0.097 & Effect is not early \\
Full-residual patching at cue token & cue token & 0.032 & Not where the cue is \\
Norm-matched random direction ($\times 20$) & L16 & 0.021 & Specific on 20/20 families \\
Unrelated-family donor & L16 & $-0.037$ & Correctly null \\
\midrule
\multicolumn{4}{@{}l}{\textit{Other architectures}} \\
Llama-3.1-8B multi-head knockout & L21 & 1.748 & CI $[-3.461, +6.653]$; spans zero \\
Llama-3.1-8B single-head knockout & L18 & 0.358 & Weak \\
DeepSeek-R1-Distill-Qwen-32B multi-head knockout & L44 & 0.086 & No effect \\
\bottomrule
\end{tabularx}
\end{table}

\begin{figure}[h]
\centering
\includegraphics[width=0.6\linewidth]{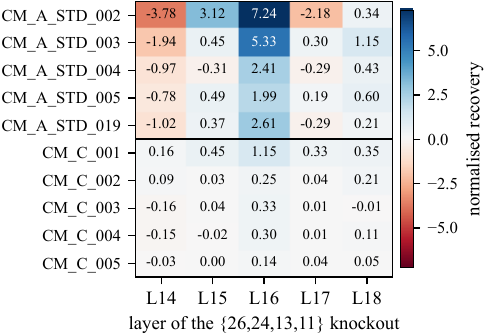}\hfill
\includegraphics[width=0.385\linewidth]{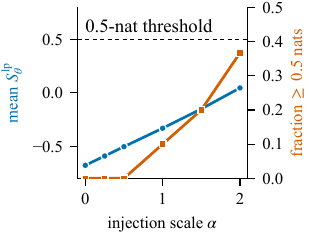}
\caption{\textbf{Left:} the same four-head knockout applied to ten held-out families it was not
selected on, five Cue A (above the line) and five Cue C (below), by layer. Every Cue A family recovers
above 0.5 at layer 16 and only one Cue C family does (the other four all recover below 0.35), so the
2.177 panel mean is carried by Cue A.
\textbf{Right:} direction-injection dose response on 30 stable families: the mean $\Slp$ rises
monotonically with $\alpha$ but the fraction crossing the threshold reaches only 0.367 at
$\alpha = 2$, twice the probe's own scale.}
\label{fig:app-injection}
\end{figure}

\paragraph{Sufficiency in full.} Direction injection on 30 stable families at
$\alpha \in \{0, 0.25, 0.5, 1.0, 1.5, 2.0\}$ gives fractions crossing the 0.5-nat threshold of
0.000, 0.000, 0.000, 0.100, 0.200 and 0.367. The stable-family mean $\Slp$ rises from $-0.678$
to $+0.046$, while five high-sensitivity controls stay flat between 10.15 and 10.34 nats
(Figure~\ref{fig:app-injection}). The recorded verdict is no sufficiency.

\paragraph{Later Llama sweeps, exploratory.} After the layer-21 result in Table~\ref{tab:causal}, a
32-head single-head sweep of Llama at layers 19 to 21 found head 2 at layer 21 with mean recovery
1.847. Combinations built on head 2 reached 2.619, with 15 of 20 families above 0.5. The experiment
registry records these runs as exploratory with incomplete provenance and not paper-eligible, so no
claim in this paper rests on them. They are noted so that the Llama null is not read as the end of
the story.

\section{Cross-Model Behavioural Comparison}\label{app:crossmodel}

\paragraph{DeepSeek against Qwen.} DeepSeek-R1-Distill-Qwen-32B
is sensitive on 27.1\% of families against Qwen's 50.0\%, with mean $\Slp$ 0.700 against 1.401 and
median 0.188 against 0.531. The difference is significant under McNemar's test, which asks whether
the two models disagree on the sensitive/stable label more often in one direction than the other on
the same families ($\chi^2 = 21.35$, $p = 3.8 \times 10^{-6}$). It is also significant under a
Wilcoxon signed-rank test, which compares the continuous statistic family by family without assuming
a normal distribution ($W = 2090.5$, $p = 4.0 \times 10^{-8}$). A $p$-value is the probability of a
difference at least this large arising
by chance. The suppression is selective: 42 families are Qwen-only
sensitive, 28 are sensitive for both, and 9 are DeepSeek-only. Fisher's exact test shows the
asymmetry directly: 36 of 53 Cue B positives are suppressed against 0 of 5 Cue C positives
($p = 5.7 \times 10^{-3}$). This is more than chance allows for a test of whether the suppressed
proportion differs between cue types. DeepSeek shows less unit-matched distractor confusion
than Qwen while showing the same unit-representation effects, a second piece of evidence that what we
are calling sensitivity is more than one phenomenon. $\Slp$ is scored by teacher forcing directly at
the prompt boundary, before any reasoning trace, and DeepSeek's reasoning-trace behaviour at that
position was never confirmed (Appendix~\ref{app:instr}). Whether this pattern reflects a completed
reasoning chain or the model's prompt-boundary behaviour alone is not established here.

\paragraph{Variable-renaming expansion.} A separate expansion of 10 families with renamed variables
gives a 0.9 positive rate with mean $\Slp$ 3.74, so sensitivity survives renaming the variables in a
family that already showed it.

\paragraph{Cross-model probe transfer.} Transfer is weaker than within-model performance in every
direction: Qwen to Llama 0.637, Llama to Qwen 0.606, Qwen to DeepSeek 0.725, DeepSeek to Qwen 0.633,
against within-model values of 0.760 and 0.703.

\section{Additional Analyses}\label{app:additional}

These analyses refine the picture in Section~\ref{sec:results}.

\textbf{(a)} \emph{Cue-specific probe directions.} Cue A and Cue B probe weight vectors at layer 16
have cosine similarity 0.018, with no layer above 0.03. A probe trained on one cue type scores
0.451 (A to B) and 0.511 (B to A) on the other. Together with the shared causal heads of
Section~\ref{sec:causal}, this indicates that sensitivity has cue-specific readouts over shared
components. \textbf{(b)} \emph{Head interaction.} Knocking out head 11 leaves head 15's contribution
to the output score unchanged (0.0\% over 40 rows), so the two act independently.
\textbf{(c)} \emph{Extraction sites.} A 7168-dimensional probe over four sites reaches 0.688. The single
layer-18 residual site at 0.760 is the most informative. \textbf{(d)} \emph{Domain transfer.} Holding
out whole physics domains gives 0.750, against a mean of 0.734 when labels are shuffled 1{,}000 times
($p = 0.368$), so domain transfer is not separable from chance at this sample size. Domain is a
coarser grouping than the 34 template clusters in Appendix~\ref{app:integrity}, so this result does
not rule out leakage at the template-cluster level.
\textbf{(e)} \emph{Proximity and the causal effect.} Numerical proximity correlates with the
behavioural label (Pearson 0.074 over 100 families, with the strongest effects in the near-match
tail) and not with head-11 recovery ($-0.039$). \textbf{(f)} \emph{Two sufficiency tests.} Direction
injection raises $\Slp$ steadily and brings 36.7\% of stable families over the threshold at twice the
probe's scale (Figure~\ref{fig:app-injection}). Cross-family activation donation from 5 sensitive
donors into 20 stable targets raises $\Slp$ by at least 0.3 nats on 86 of 100 pairs, with a mean
increase of 3.56 nats. Donation carries more than the single probe direction. \textbf{(g)}
\emph{Renamed-donor controls.} The donor-on-renamed specificity controls produced no evaluable rows
because of a donor-selection defect. This control is unmeasured, and the specific fraction of 0.681
rests on the stable-donor and same-answer-donor decompositions. \textbf{(h)} \emph{Frozen-probe
transfer, by cue type.} The layer-16 probe frozen after training (Section~\ref{sec:monitor}) reaches
AUROC 0.820 (AUPRC 0.907) pooled over the 15 families built after training, 5 Cue B and 10 Cue C.
Restricted to the five Cue B families, AUROC is 0.667 (AUPRC 0.750). Restricted to the ten Cue C
families, AUROC is 0.6875 (AUPRC 0.907). The pooled figure exceeds either subgroup because the
probe's scores cluster high on Cue C families and low on Cue B families. Cue B's true labels are,
meanwhile, majority-negative (3 of 5) and Cue C's are majority-positive (8 of 10). This alignment of a
between-group score offset with a between-group base-rate skew inflates the pooled AUROC even though
within-subgroup ranking is weak. Both subgroups are too small (n=5, n=10) to resolve precise AUROC
values.

\section{Failure Taxonomy}\label{app:failures}

The monitor's confident errors cluster into four structural categories, which bears on how it would be deployed.

\textbf{(a)} \emph{Dangerous template, clean rejection.} \texttt{CM\_B\_UM\_018}, an Ohm's-law family
with a unit-matched isolated branch current, is scored 0.966 with a measured $\Slp$ of exactly 0.
\texttt{CM\_B\_UM\_041}, momentum to velocity with a disconnected 6.0 m/s distractor, is scored 0.998
with $\Slp = 0.203$. Both instantiate the geometry of a sensitive family without the numerical
proximity that triggers the behaviour.

\textbf{(b)} \emph{Right shell, distant distractor.} \texttt{CM\_B\_UM\_035} and
\texttt{CM\_B\_UM\_053}, pendulum and spring-mass families with frequency distractors at 0.5 and 0.8
Hz, are scored high and remain behaviourally stable.

\textbf{(c)} \emph{Weak positives underestimated.} \texttt{CM\_B\_UM\_015} ($\Slp = 0.578$, scored
0.125) and \texttt{CM\_B\_UM\_027} ($\Slp = 0.719$, scored 0.131) clear the threshold with cue values
that are not especially close to the answer.

\textbf{(d)} \emph{Strong positives with dispersed signatures.} \texttt{CM\_B\_UM\_009}
($\Slp = 4.375$, scored 0.309), \texttt{CM\_B\_004} ($\Slp = 2.672$, scored 0.213) and
\texttt{CM\_A\_STD\_019} ($\Slp = 3.156$, scored 0.365) are unambiguous positives the layer-18 mean
feature underestimates.

Categories (a) and (b) are the same error viewed at different confidence levels, and they say the
probe reads template class. Category (d) says a single mean feature does not capture every route to
high sensitivity. Together they motivate the numerically aware per-rendering representation named in
Section~\ref{sec:discussion}.

\section{Scope of the Claims}\label{app:scope}

All results are on open-weight models between 7B and 32B parameters, on text-only
physics problems in three domains with four renderings per family, under teacher-forced scoring of
a single reference answer. The parser is part of the measurement definition, and while $\Slp$ is
parser-independent by construction, $\Sflip$ inherits its coverage. We do not claim these findings
extend to multimodal physics, closed-weight models, or traces produced under sampling. Within these
bounds the comparisons are internally consistent: every predictor in Figure~\ref{fig:incr} is scored
on the same 135 families under the same leave-one-family-out split, and every interval resamples the
same unit.

\section{Benchmark Integrity Audits}\label{app:integrity}

\paragraph{Verifier mutation testing.} Over 30 families, 330 mutation records were generated by
applying a fixed set of mutation operators. 300 were verified against expectation. At the operator
level, every positive operator was accepted at rate 1.0 (30/30 algebraically equivalent expression
rewrites) and every negative operator was rejected at rate 1.0 (30/30 cue-leakage-after-assumption-
removal mutations). Three operators whose names did not describe what they executed were removed,
and the audit records the removal.

\paragraph{Balance.} No single surface feature predicts the label: orientation-independent AUC is
0.503 for prompt length, 0.509 for token count, 0.557 for number count. Domain shows no association
with the label (Cram\'er's $V = 0.092$, $p = 0.551$). Cue type ($V = 0.251$, $p = 0.012$) and template
cluster ($V = 0.382$, $p = 0.009$) do associate, which is expected given Table~\ref{tab:bench} and is
the reason every evaluation in this paper holds out whole families. A
family-row stratified classifier over surface features alone reaches 0.612 with refit permutation
$p = 0.015$, below every sensitivity-trained probe in Table~\ref{tab:probes} except two. These are the
embedding-layer probe, which reads the raw input representation before any transformer processing,
and the prompt-condition probe, which is trained on the cue value (a Level-1 target), not on
sensitivity. Fifteen families
without authoritative labels are excluded. An earlier version of this
audit coerced them to negative and is superseded.

\paragraph{Leakage.} A structure-first audit over the full untruncated population of 620 prompts and
155 families finds 0 cross-split exact duplicates. Of 24{,}354 candidate pairs after blocking,
21{,}072 are resolved as shared prompt scaffold and 930 are within-family expected pairs, leaving
2{,}352 unresolved cross-family rendering pairs, corresponding to 147 family pairs in 34 clusters and
touching 110 families. Active detectors are exact prompt hash, number-masked prompt hash, normalised
symbolic equation hash, cue template hash, variable-renaming invariant structure hash, and numeric
instantiation sibling detection. Lexical similarity is used only as a tiebreaker, because the shared
prompt scaffold makes a lexical threshold near-universal, a defect that affected an earlier version
of this audit, which is superseded here. Semantic template similarity was unavailable without a local embedding method and is
recorded as not run. Template clusters are a coarser grouping than family identity. The primary
evaluation (Section~\ref{sec:setup}) holds out whole families but not whole clusters, so this audit
does not rule out a held-out family sharing cluster-level structure with a family in its training
fold.

\paragraph{Provenance.} 182 science runs carry real scheduler job identities, with registry coverage
0.703 and strict coverage 0.451 against that denominator. All critical artefacts carry validated
resolution records, and all ten tracked claims cite an authoritative artefact. Every headline number
in this paper is registered against a named artefact with an allowed and prohibited wording note, and
the wording used in Sections~\ref{sec:behaviour} through~\ref{sec:limitations} conforms to those
notes.

\paragraph{Human validation status.} Dual independent validation of cue irrelevance is specified, with
an annotation schema, qualification test, calibration bank, validator handbook and stratified pilot
selection all committed. The pilot has not been run, and until it is, the benchmark's cue-irrelevance
claims rest on symbolic certification alone. Symbolic certification is the stronger of the two
guarantees for the invariance itself, since it proves the answer cannot change. Human validation
addresses the separate question of whether a cue is \emph{plausibly} irrelevant to a physicist reading
the prompt, which certification does not establish.

\end{document}